\documentclass[conference]{IEEEtran}
\IEEEoverridecommandlockouts
\usepackage{cite}
\usepackage{amsmath,amssymb,amsfonts}
\usepackage{algorithmic}
\usepackage{graphicx}
\usepackage{textcomp}
\usepackage{xcolor}

\usepackage{caption} 
\usepackage{subcaption} 
\usepackage{listings}
\usepackage{tikz} 
\usepackage{booktabs}
\usepackage[hidelinks]{hyperref}

\def\BibTeX{{\rm B\kern-.05em{\sc i\kern-.025em b}\kern-.08em
    T\kern-.1667em\lower.7ex\hbox{E}\kern-.125emX}}

\usepackage{fancyhdr}
\makeatletter
\let\old@headrule\headrule
\renewcommand{\headrule}{\if@fancyplain\let\headrulewidth\plainheadrulewidth\fi\old@headrule}
\makeatother

\renewcommand{\headrulewidth}{0pt}

\makeatletter
\def\ps@IEEEtitlepagestyle{%
  \def\@oddhead{\mbox{}\scriptsize\rightmark \hfil}%
  \def\@evenhead{\scriptsize\thepage \hfil \leftmark\mbox{}}%
  \def\@oddfoot{\hfil \mbox{}\parbox{5.5in}{\centering
  \footnotesize \textcopyright 2024 IEEE. Personal use of this material is permitted.
  Permission from IEEE must be obtained for all other uses, in any current or future
  media, including reprinting/republishing this material for advertising or promotional
  purposes, creating new collective works, for resale or redistribution to servers or
  lists, or reuse of any copyrighted component of this work in other works.}\hfil \mbox{}}%
  \def\@evenfoot{\mbox{}\parbox{5.5in}{\centering}\hfil \mbox{}\hfil}%
}
\makeatother

\begin{document}

\title{G-PCGRL: Procedural Graph Data Generation via Reinforcement Learning
\thanks{This research was supported by the Volkswagen Foundation (Project: Consequences of Artificial Intelligence on Urban Societies, Grant 98555).\\ \\
979-8-3503-5067-8/  24/\$31.00~\copyright2024 IEEE \hfill} 
}

\author{\IEEEauthorblockN{Florian Rupp}
\IEEEauthorblockA{\textit{Department of Computer Science} \\
\textit{University of Applied Sciences Mannheim}\\
Mannheim, Germany \\
f.rupp@hs-mannheim.de}
\and
\IEEEauthorblockN{Kai Eckert}
\IEEEauthorblockA{\textit{Department of Computer Science} \\
\textit{University of Applied Sciences Mannheim}\\
Mannheim, Germany \\
k.eckert@hs-manneim.de}
}

\IEEEpubidadjcol


\maketitle

\begin{abstract}
Graph data structures offer a versatile and powerful means to model relationships and interconnections in various domains, promising substantial advantages in data representation, analysis, and visualization. In games, graph-based data structures are omnipresent and represent, for example, game economies, skill trees or complex, branching quest lines.
With this paper, we propose G-PCGRL, a novel and controllable method for the procedural generation of graph data using reinforcement learning. Therefore, we frame this problem as manipulating a graph's adjacency matrix to fulfill a given set of constraints. Our method adapts and extends the Procedural Content Generation via Reinforcement Learning (PCGRL) framework and introduces new representations to frame the problem of graph data generation as a Markov decision process. 
We compare the performance of our method with the original PCGRL, the run time with a random search and evolutionary algorithm, and evaluate G-PCGRL on two graph data domains in games: game economies and skill trees. The results show that our method is capable of generating graph-based content quickly and reliably to support and inspire designers in the game creation process. In addition, trained models are controllable in terms of the type and number of nodes to be generated.
\end{abstract}



\section{Introduction}

Graph-based data in games is omnipresent. It represents key components such as game economies, skill trees, or complex branching quest lines. How a game \emph{feels} is heavily influenced by the design and balance of its economy, such as how resources are created, consumed, and can be transitioned to other resources. These concepts are common to many genres, including role-playing games (e.g., mana or skill development), first-person shooters (e.g., health), or simulation games (e.g., resource transitions).
Game economies, including progression systems such as skill trees, primarily create incentives and engage players in the overall gaming experience indirectly. Therefore, they can be difficult to design and game developers require a lot of manual effort in the process~\cite{schreiber_game_2021,adams_fundamentals_2014}.
For this reason, game designers use tools like Machinations\footnotemark\cite{klint_micro-machinations_2013} or are supported by works on procedural content generation (PCG) such as game levels~\cite{togelius_search-based_2011,khalifa_pcgrl_2020,awiszus_toad-gan_2020}, balancing~\cite{lara-cabrera_balance_2014,rupp_simulation_2024}, or game rules~\cite{cook_mechanic_2013}. Limited work\footnotetext{https://machinations.io/}, however, has been published about PCG for graph data. Existing works generate graph-structured representations of game content, such as economies~\cite{rogers_using_2023,rupp_geevo_2024} or narratives~\cite{alvarez_tropetwist_2022} with evolutionary algorithms.

\begin{figure}
        \centering
         \includegraphics[width=0.45\textwidth]{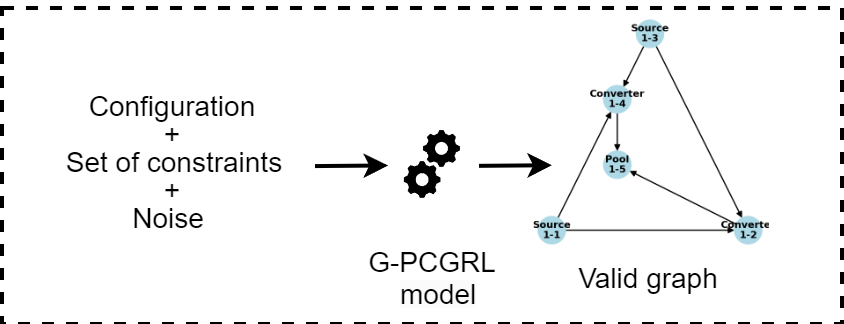}
          \label{fig:per-validity}
  \caption{Graph generation with G-PCGRL: A G-PCGRL model is controllable through a given configuration to generate a valid graph from random noise according to a set of constraints on which it has been trained.}
  \label{fig:overview}
\end{figure}

With this work, we aim to support the automatic generation of game-related graph-based structures by introducing G-PCGRL, a novel method for procedural graph data generation using reinforcement learning (RL). Once trained, an RL model can generate content quickly and is less dependent on randomness compared to evolutionary methods~\cite{khalifa_pcgrl_2020}. The generation of graph data including RL is highly researched in others fields like chemistry in the context of molecules~\cite{you_graph_2018,leguy_evomol_2020,celton_deep_2019}. To generate new realistic molecules, the work there incorporates existing data. When creating games, however, this is in most cases not applicable since there is no data to learn from.

In the games community, the procedural content generation via reinforcement learning (PCGRL)~\cite{khalifa_pcgrl_2020} framework for the generation of game levels has been introduced. In PCGRL, the RL agent learns to alter an integer-represented tile-based level -- for this work, we frame this as the adjacency matrix of a graph. By entering parameters into the initialization process, trained models can be further controlled in terms of the size and number of different node types of the graph to be generated (Fig.~\ref{fig:overview}). The code for G-PCGRL is available on Github\footnote{https://github.com/FlorianRupp/g-pcgrl}.

\newpage
Our contributions are:
\begin{itemize}
    \item G-PCGRL: A novel and controllable approach to using PCGRL to generate graph data by manipulating a graph's adjacency matrix.
    \item New graph representations and modeling approaches to frame the problem as a Markov decision process (MDP). By defining sets of constraints, the method learns how to generate valid graphs accordingly.
    \item A study to evaluate the proposed method with respect to different maximum graph sizes, number of rules, node types, different representations, computation time, and applications in two domains: game economies and skill trees.
\end{itemize}

The paper is structured as follows: We give a brief overview of related work in Section~\ref{sec:related-work} and describe the method in terms of graph representations, the rule set declaration and the reward design in Section~\ref{sec:method}. Experiments and results are presented in~\ref{sec:experiments}, followed by the discussion~(\ref{sec:discussion}) and conclusion~(\ref{sec:conclusion}).

\section{Related Work}
\label{sec:related-work}

Several approaches have been proposed for the procedural generation of graph data for games. Rogers et al. use an evolutionary algorithm (EA) to generate graph-structured game economies \cite{rogers_using_2023}. With an additional user study, the authors show that this method can generate economies of varying complexities for a simulation game. 
Rupp et al. present GEEvo, a framework for generating and balancing graph-structured game economies~\cite{rupp_geevo_2024}. The graphs are generated by random mutations applied to their edge list.
However, EAs rely heavily on randomness, whereas with RL models, once trained, can generate content quickly and reliably. To the best of our knowledge, no work has been published yet on generating graph data for games with RL.

Besides search-based approaches, graph rewriting systems, such as graph grammars, are widely used to generate graph-based structures. Graph grammars provide a flexible way to model complex structures and are used in the context of games to generate levels, as in platformers or dungeon layouts ~\cite{valls-vargas_graph_2017,hauck_automatic_2020,gutierrez_generative_2020}, rules~\cite{cook_mechanic_2013}, or graphs as the basis for narratives~\cite{alvarez_tropetwist_2022}. Whereas graph grammars apply rules iteratively, this work focuses on the learning of a set of constraints in a general way from which a graph can be constructed.

The generation of graph data is not limited to game-related content. It is also widely used in the field of chemistry, where a common problem domain is the optimization of the properties of molecular graphenes.
Celton et al. provide a detailed survey on how to apply deep learning for graph-based molecular design\cite{celton_deep_2019}, Leguy et al. generate molecules with an EA~\cite{leguy_evomol_2020}, and Kwon et al. apply an EA combined with recurrent neural networks~\cite{kwon_evolutionary_2021}.  Zhou et al. formulated the problem as an MDP to apply RL~\cite{zhou_optimization_2019}. The molecule graph generation is a sequential task where the agent's action space is to add atoms and to create or remove bonds. In this work, we formulate the problem differently and manipulate the graph's adjacency matrix as opposed to a sequential task.
You et al. combine a graph neural network (GNN) with RL to generate molecules with an optimized drug-likeliness~\cite{you_graph_2018}. The GNN is used to create node embeddings based on existing data, the molecules are then generated with RL, where the agent's action space is to connect atoms (nodes) and subgraphs through link prediction.
The key distinction for this work from chemistry-related graph data generation is that molecular graph generation uses real-world data. The constraints for generating new molecules are derived or learned from this data. In contrast, this work addresses a scenario where no data exists, only abstract concepts.

With G-PCGRL, we focus on procedural content generation for graph data generation, but the term PCG is also used in other applications such as the generation of narratives or textures. PCG methods range from dedicated algorithms~\cite{mojang_mincraft_2011}, to search-based methods~\cite{togelius_search-based_2011,rogers_using_2023}, to the use of machine learning~\cite{summerville_procedural_2018} and deep learning~\cite{liu_deep_2021,awiszus_toad-gan_2020}. Recently, Large Language Models (LLMs) have also been shown to generate prompt-controllable content for Super Mario levels~\cite{sudhakaran_mariogpt_2023}.
Once trained, machine learning-based methods can quickly generate content on demand, but they rely on the existence of content from which a model can be trained. As a result, these methods are usually not efficiently applicable to the creation of new games.

In the games community, RL has been widely applied to play games~\cite{vinyals_grandmaster_2019,silver_general_2018}, but recently also for PCG tasks~\cite{khalifa_pcgrl_2020,gisslen_adversarial_2021} as well. In this work, we use and adapt PCGRL~\cite{khalifa_pcgrl_2020} to generate graph data. Compared to other deep learning-based PCG approaches for the puzzle game Sokoban, Zakaria et al. found that PCGRL produced higher quality results~\cite{zakaria_procedural_2022}.
There are already a few methods that have adapted PCGRL. Earle et al. introduce controllable content generators~\cite{earle_learning_2021}, where users control the content with additional constraints, such as the number of players. In contrast, controllability in G-PCGRL is provided by the configuration of the initial matrix. Rupp et al. propose to use PCGRL as a fine-tuning procedure for existing content using swapping representations and demonstrate this on a game balancing task~\cite{rupp_balancing_2023,rupp_simulation_2024}. For G-PCGRL, however, these swapping representations are not applicable as they require already existing content.
Jiang et al. adapt PCGRL for 3D environments~\cite{jiang_learning_2022-4}.

\section{Method}
\label{sec:method}

\label{sec:background}

In this work, we extend the PCGRL \cite{khalifa_pcgrl_2020} framework for the procedural generation of graph-based data.
In PCGRL, PCG is formulated as a sequential decision-making task to maximize a given reward function, where semantic constraints can be expressed and thus, no training data is needed. To apply RL, the PCG problem is formulated as an MDP. Therefore, PCGRL introduces three different MDP representations for level generation: \textit{narrow}, \textit{turtle}, and \textit{wide}. The representations differ in the way they model the observation and action space for the agent. Following the ideas of the original paper, we introduce two new graph representations based on the narrow and wide representations. We do not work with the PCGRL turtle representation due to its limited ability to manipulate the matrix equally.


\subsection{Graph representations}
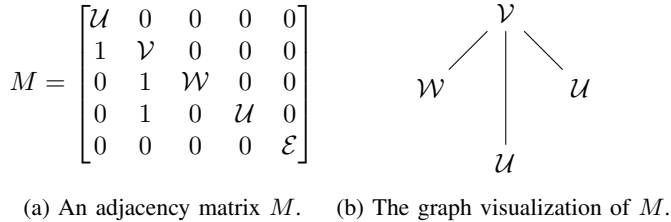
\begin{figure}
    \begin{subfigure}{0.25\textwidth}
        \[
        M=
          \begin{bmatrix}
            \mathcal{U} & 0 & 0 & 0 & 0\\
            1 & \mathcal{V} & 0 & 0 & 0\\
            0 & 1 & \mathcal{W} & 0 & 0\\
            0 & 1 & 0 & \mathcal{U} & 0\\
            0 & 0 & 0 & 0 & \mathcal{E}\\
          \end{bmatrix}
        \]
       \centering
      \caption{An adjacency matrix $M$.}
      \label{fig:adj-matrix}
    \end{subfigure}%
    \begin{subfigure}{0.25\textwidth}
        \centering
      \begin{tikzpicture}[scale=0.5]
          \draw
            (0.0:2) node (U 1){$\mathcal{U}$}
            (90.0:2) node (V 2){$\mathcal{V}$}
            (180.0:2) node (W 3){$\mathcal{W}$}
            (270.0:2) node (U 4){$\mathcal{U}$};
          \begin{scope}[-]
            \draw (U 1) to (V 2);
            \draw (V 2) to (W 3);
            \draw (U 4) to (V 2);
          \end{scope}
        \end{tikzpicture}
        \caption{The graph visualization of $M$.}
        \label{fig:graph-example}
    \end{subfigure}
    \caption{Overview of the representation of a graph (right) as an extended adjacency matrix $M$ (left). The nodes are represented on the matrix's diagonal and are encoded depending on the node type. Edges between nodes are represented as 1, 0 for no connection. 
    } 
    \label{fig:adj-matrix-overview}
\end{figure}

The RL agent in PCGRL modifies an integer matrix representing a tile-based game level. For G-PCGRL, this matrix is considered to be the adjacency matrix of a graph (Fig.~\ref{fig:adj-matrix-overview}). A simple example of such an adjacency matrix $M$ is given in Fig.~\ref{fig:adj-matrix}. For this research, we extend the conventional concept of an adjacency matrix to represent the node types on the matrix's diagonal. In this example, we encode different types of nodes, such as $\mathcal{U}$, $\mathcal{V}$, and $\mathcal{W}$. The information as to whether two nodes are connected by an edge is represented by a 1, 0 stands for no connection. To allow controllability in terms of also generating graphs smaller than the size of $M$, we pad $M$ with additional symbols of type $\mathcal{E}$ (empty).
Since we are dealing with undirected graphs here and infer the direction from the domain (cf.~ Section~\ref{sec:declaration-constraints}), no information is yet assigned to the upper right area of the adjacency matrix, resulting in a triangular shape of the action space.

To formulate the problem of graph data generation as an MDP, we need to define the 4-tuple ($S,A,P,R$), where $S$ represents the set of states and $A$ the set of actions. $P$ is the state transition probability function and $R$ is the reward function, defining the dynamics and rewards of the environment.
In PCGRL, $S$ (also called the observation space) and $A$ are modeled with different \emph{representations}. In this work, we introduce the \emph{graph-narrow} and \emph{graph-wide} representations, both inspired by the corresponding ones in the PCGRL paper.

For all representations, the action space is the prediction of whether two nodes are connected or not, in other words the area below the adjacency matrix's diagonal. The diagonal cannot be modified by the agent, it is randomly created or can be predefined to allow controllability.
Since all values in the matrix represent labels, we apply a One-Hot-Encoding in all representations as the final transformation for the observations as in PCGRL.

\subsubsection{Graph-narrow representation}

Like in the PCGRL narrow representation, we implement the position selection through the environment. Since it is important that the agent is given the opportunity to potentially modify any position of the adjacency matrix, we use the narrow sequential mode. The episode terminates when either a complete iteration over the action space has been performed, a maximum number of changes or iteration has been exceeded, or the graph is valid (cf.~\ref{sec:graph-valid}).

In the context of a graph, a position represents the state of an edge between two nodes. Using this representation, the action space is very small and consists of only two actions: toggle the state of an edge, or not. To encode the information about which edge between which nodes has been selected for modification, an appropriate observation is required. Therefore, we model this as two vectors, each representing a selected node and its connections. This is comparable to the cropping approach in PCGRL narrow.
An example is shown in Fig.~\ref{fig:graph-narrow}. In this example, the position marked with a black frame $(0,2)$ is selected by the environment; the resulting observation space is marked with red frames. Here it is the two vectors representing the nodes $\mathcal{U}$ and $\mathcal{W}$ with all their edges.
The advantage of the graph-narrow representation is its small action space, however, its cropped observation space and the limited selection mechanism can be a disadvantage, as there is less information available to the agent.

\begin{figure}[!t]
  \begin{subfigure}[b]{0.5\textwidth}
      \centering
      \includegraphics[width=1.1in]{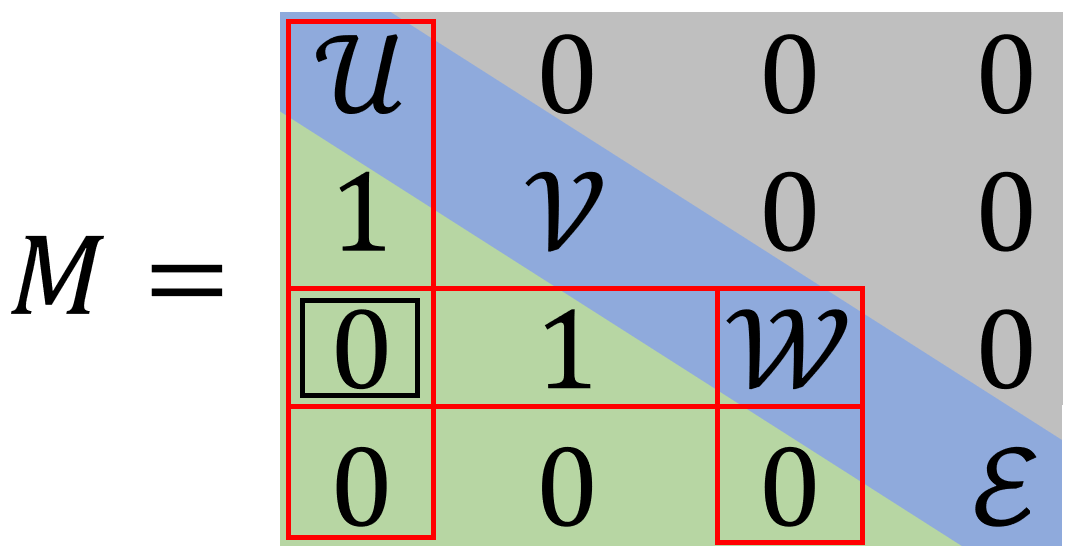}
      \caption{The adjacency matrix in the graph-narrow representation: A position, i. e. an edge (black frame), is selected by the environment. The observation consists of the two affected nodes and their connections (red frames). 
      }
      \label{fig:graph-narrow}
  \end{subfigure} %
  \hspace{0.5cm}
    \begin{subfigure}[b]{0.5\textwidth}
      \centering
      \includegraphics[width=1.1in]{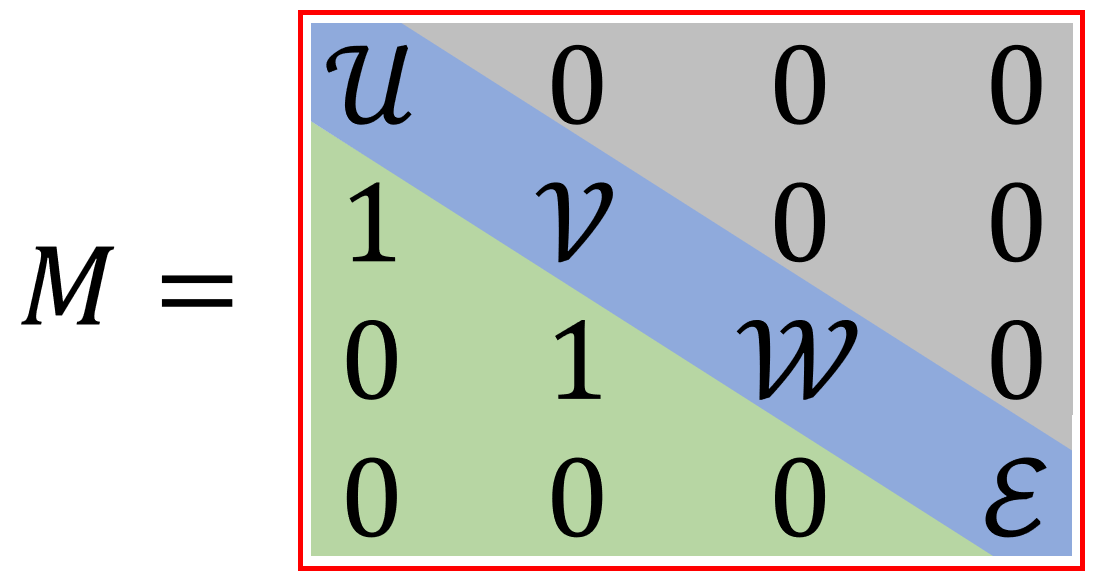}
      \caption{The adjacency matrix in the graph-wide representation: the agent is free to select any position in the matrix that is in its action space (green). The observation space is the entire matrix (red frame).\newline}
      \label{fig:graph-narrow}
  \end{subfigure}
  \caption{Overview of the action and observation spaces for the graph-narrow (left) and graph-wide (right) representation. The action space is highlighted with a green area, the observation space with red frames. The diagonal (blue) is static for both representations and can be configured at initialization or is generated randomly. Since we are only dealing with undirected graphs so far, the gray area is not used, but is important for defining the shape of a square matrix.}
  \vspace{-4mm}
\end{figure}

\subsubsection{Graph-wide representation}

As in the PCGRL wide representation, the agent is given full control in the graph-wide representation. For each time step, the agent predicts a position on the adjacency matrix, i.e. an edge, and whether this edge state should be switched or not. Full observation of the adjacency matrix provides the agent with maximum information, but significantly increases the action and observation space. The size of the action space $A$ for a graph with $n$ nodes can be calculated using the triangular formula (Eq.~\ref{eq:triangular}) with $n-1$, since the diagonal is not part of $A$.
The RL agent in G-PCGRL predicts an action $a$ as a discrete value. To map this action to a 2D position in the matrix, we use the discriminant of the square polynomial (cf. triangular root, Eq.~\ref{eq:triangular-root}) and round it up. The y-position is then calculated with $triangRoot(a)$ and the x-position is calculated from the resulting y-position with~$a-triang(y-1)-1$. 
The episode ends when the graph is either valid (see~\ref{sec:graph-valid}), or a maximum number of changes or iterations is exceeded.

\begin{equation}
\label{eq:triangular}
triang(n) = \frac{n \cdot (n + 1)} {2}
\end{equation}

\begin{equation}
\label{eq:triangular-root}
triangRoot(x) = -\lceil \frac{ -1 + \sqrt{1 + 8x}} {2} \rceil
\end{equation}


\subsection{Declaration of a set of constraints}
\label{sec:declaration-constraints}
With a set of constraints, abstract concepts can be defined which rules the graph content must follow, thereby defining the search space for the RL agent. The constraints determine which node types must be connected and, conversely, which must not be connected.
To keep the syntax simple, human-readable, easily editable, and extendable, we choose a JSON-like syntax, as described in the example in Fig.~\ref{lst-example-constraints}. For each node type, we declare a list of node types, each of which must be connected to a node of that type with a direct edge. The number of connected valid node types can be $>= 1$. By default, direct connections to node types that are not included in the set of allowed edges of a node type are not allowed. Although only this type of constraints is supported, it can also be considered as a simple and lightweight ontology.

Depending on the node size of a graph and the number of nodes of the same type, a set of constraints can allow for multiple different graphs. Fig.~\ref{fig:graph-example} shows an example of a valid graph of the set of constraints in Fig.~\ref{lst-example-symbolic}. In a literal sense, these constraints require that a node of the symbolic type $\mathcal{U}$ must be connected to at least one node of type $\mathcal{V}$, $\mathcal{V}$ must be connected to at least one $\mathcal{U}$ and $\mathcal{W}$, and $\mathcal{W}$ must be connected to at least one $\mathcal{V}$.
A more concrete use case for a game economy setting is given in Fig.~\ref{lst-example-economy}, where we substitute the symbolic node types with entities of a typical game economy setting. We stick to the node naming for game economies established by Klint et al.~\cite{klint_micro-machinations_2013} which has also been used along with a set of constraints in~\cite{rupp_geevo_2024}.

$\mathcal{U}$ represents a node of type \texttt{Source}, which is an entity where resources are added to the economy, such as a mine or a tree that can be farmed. $\mathcal{V}$ is a \texttt{Converter} where one or more resources from \texttt{Source} nodes are converted into a new resource (e.g., an item). This abstract design logic is thereby enforced by the set of constraints.
The newly transitioned resources are held then in a \texttt{Pool} node. In this narrative setting, we can infer the directions from the domain: \texttt{Source} always is directed to \texttt{Converter} and \texttt{Converter} to \texttt{Pool}. 
In the game Minecraft, for instance, the process of crafting (Converter) torches (Pool) from wood (Source) and coal (Source) can be illustrated this way. This concept is not unique to Minecraft and can be found in many other games~\cite{schreiber_game_2021}.


\begin{figure}[htb]
     \begin{subfigure}[b]{0.45\textwidth}
        \centering
        \begin{lstlisting}[frame=tb, language=Python, linewidth=1\textwidth, xleftmargin=0.01\textwidth, xrightmargin=0.01\textwidth,escapechar=§]
            §$\mathcal{U}$§: §$\hspace{0.35em}$§[§$\mathcal{V}$§]
            §$\mathcal{V}$§: §$\hspace{0.375em}$§[§$\mathcal{U}$§, §$\mathcal{W}$§]
            §$\mathcal{W}$§: [§$\mathcal{V}$§]
        \end{lstlisting}
          \caption{Notation with generalized symbols.}
          \label{lst-example-symbolic}
     \end{subfigure}
     \begin{subfigure}[b]{0.45\textwidth}
     \hspace{0.5cm}
         \centering
        \begin{lstlisting}[frame=tb, language=Python, linewidth=1\textwidth, xleftmargin=0.01\textwidth, xrightmargin=0.01\textwidth,escapechar=§]
        Source:    [Converter]
        Converter: [Source, Pool]
        Pool:      [Converter]
        \end{lstlisting}
        \caption{Notation for a game economy setting (cf.~\cite{klint_micro-machinations_2013}).}
         \label{lst-example-economy}
         
     \end{subfigure}
     
  \caption{An example set of constraints for graph data generation, consisting of four rules and three different node types. The same constraints are written with generalized symbols (left) and in a game economy use case (right). G-PCGRL can be trained with an arbitrary set of constraints, we will experiment with several ones that are listed in the appendix (Fig.~\ref{fig:app-constraints}).}
  \label{lst-example-constraints}
\end{figure}

\subsection{Reward design and graph validation}
Appropriate reward design is crucial for the successful use of RL. As in PCGRL, we use an intermediate reward to reward the agent at each time step $t$ according to the change it has made from $t-1$ to the validity of the graph $g_t$ in the context of the given set of constraints with a validity function $v$. We consider a graph to be valid if all given constraints hold for all nodes. 
If the action removed an incorrect edge or created a missing one, the validity reward of $v(g_t, g_{t-1})$ is positive, otherwise it is negative. There is no reward (value 0) if the agent's action did not affect the validity.
Since in both representations only the creation or removal of edges between nodes can be performed, there are exactly these five possible states for rewarding.
If $g_t$ is valid, we add an additional $\alpha$ to reward the agent. If the graph is invalid, $\alpha$ is zero. The reward $r_t$ at a time step is then $r_t=v(g_t, g_{t-1}) + \alpha$.


\label{sec:graph-valid}

\section{Experiments and Results}
\label{sec:experiments}

\begin{figure*}
          \centering
         \includegraphics[width=0.92\textwidth]{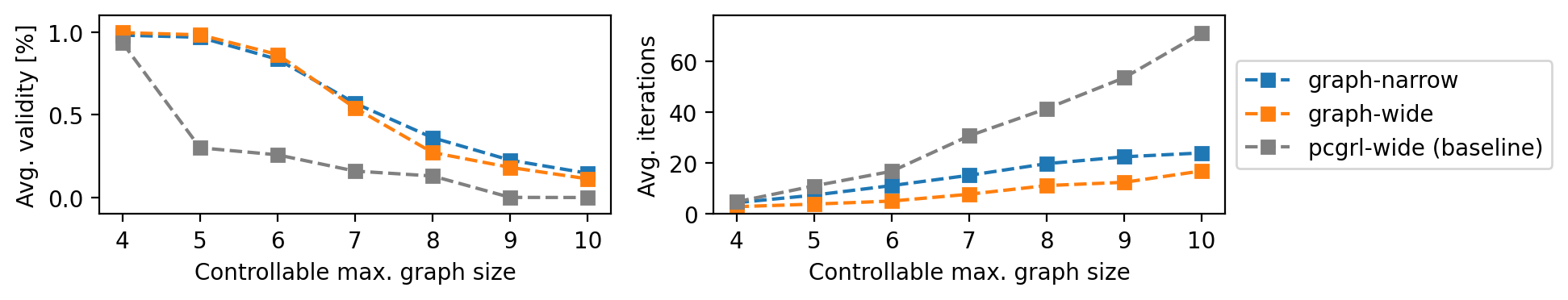}
          \label{fig:per-validity}
  \caption{The overall performance of the controllable G-PCGRL models compared to the different representations based on the maximum controllable graph size from a sample of 500 each. Performance is measured as the percentage of valid graphs generated (left). The average iterations per maximum graph size are shown in the right subplot. We compare both newly introduced representations to the original PCGRL-wide as a baseline (gray).}
  \label{fig:performance}
    \vspace{-2mm}
\end{figure*}

To evaluate our method, we train multiple models with different configurations for the graph-narrow and graph-wide representations. The evaluation is done in several steps: First, we describe the experimental setup in Section~\ref{sec:exp-setup}. The overall performance is evaluated in Section~\ref{sec:performance}.
Finally, we demonstrate the ability of G-PCGRL to generate appropriate graphs for two different game use cases: the generation of economy structures and skill trees (Section~\ref{sec:demo}).

\subsection{Experimental setup}
\label{sec:exp-setup}

For all representations we test different configurations like different sets of constraints (e.g., number of contained constraints), the number of different node types, and the graph size (number of nodes). The latter refers to a graph size up to a maximum size, as all models are trained to be controllable to also generate arbitrarily smaller graphs.
An overview of the different parameters and their values is given in Table~\ref{tab:exp-setup}.
This results in 105 trained models, which will be compared and evaluated in this section. All models are trained on a specific set of constraints but are controllable in terms of the size of the graph and the exact number of node types. To train the models in a controllable manner, we ensure that all possible configurations are sampled uniformly during training.
To compute metrics for comparison, we generate a sample of 500 graphs from each model.

For all experiments, we use Proximal Policy Optimization (PPO)~\cite{schulman_proximal_2017} with a multi-layer perceptron architecture for the feature extractor and the value function. We configure both with three fully connected layers of sizes 128, 256, and 128. The graph-narrow representation models are all trained for 500k steps, resulting in 400 policy updates. All graph-wide and pcgrl-wide models are trained for 1.5 million steps (1200 policy updates) due to preliminary investigations have shown that the larger action space requires more training steps to converge.

\begin{table}
    \centering
\caption{Overview of parameters for experimental setup.}
\label{tab:exp-setup}
    \begin{tabular}{ll} \toprule
         \textbf{Parameter} &  \textbf{Values} \\ \cmidrule(lr){1-1} \cmidrule(lr){2-2}
         Representations &  graph-narrow, graph-wide, pcgrl-wide\\ 
         Graph sizes up to &  4, 5, 6, 7, 8, 9, 10\\ 
         Number of node types&  2, 3\\ 
 Sets of constraints & Set1 (Fig.~\ref{lst-example-constraints}), Set2*, Set3*, Set4*, Set5*\\ \bottomrule
 \vspace{-2mm} \\
  \multicolumn{2}{l}{$^{\mathrm{*}}$The sets of constraints 2-5 are listed in the appendix in Fig.~\ref{fig:app-constraints}.}
  \vspace{-4mm}
\end{tabular}
\end{table}

     

\subsection{Performance overall}
\label{sec:performance}

We measure the overall performance of a G-PCGRL model by estimating the average proportion of validly generated graphs from the sample of 500 per model. The average validity for all models per representation and per maximum graph size is shown in Fig.~\ref{fig:performance} (left). Both graph representations show a similar performance, but the pcgrl-wide as a baseline is significantly worse. For smaller graph sizes, the mean validities for both graph representations are at their maximum, however, as the graph size increases, the performance gets worse.
Figure~\ref{fig:performance} (right) compares the average iterations of all models per representation and per maximum graph size. Iterations are the number of actions taken by an agent within an episode before termination. An episode is over when the graph is valid, or when a maximum number of changes or total iterations is exceeded. The larger the graph size, the more iterations were required by the agents in both representations. The graph-wide representation, however, requires fewer iterations than the graph-narrow representation. The pcgrl-wide baseline requires comparatively many more iterations than the graph representations.


\subsection{Execution time}
A major advantage of using RL for PCG is its remarkably fast inference speed. We evaluate this by contrasting it with a random search and an EA. The experimental setup is as follows: each method is given the same graph configuration and a set of constraints to generate a graph. Execution time\footnote{We use a 2.6 GHz AMD EPYC family 23 model 1 processor for all experiments.} is measured as the time required for each method to produce a valid graph from the given combination of configuration and set of constraints. For each method, set of constraints, and graph size, 100 graphs are generated.

\begin{table}
    \centering
    \begin{tabular}{cc|ccc}
        \toprule
        \textbf{Constraints} & \textbf{Size} &  \textbf{G-PCGRL} & \textbf{Evol. algorithm}  & \textbf{Random search} \\ 
        \midrule
        Set 1 & 5 &  \textbf{6.6} &  42.2 & 96.6\\
        & 6 & \textbf{9.4} & 59.7 & 378.1\\
        & 7 & \textbf{5.6} & 109.1 & 2509.7 \\ \midrule
        Set 2& 5 & \textbf{7.6} & 49.7 & 457.3\\
        & 6 & \textbf{10.8} & 107.6 & 6811.9 \\
        & 7 & \textbf{5.4} & 143.1 & 16112.2 \\ \midrule
        Set 3& 5 & \textbf{6.3} & 26.1 & 37.7 \\
        & 6 & \textbf{6.3} & 45.4 & 147.1 \\
        & 7 & \textbf{3.5} & 67.3 & 750.8 \\ \midrule
        Set 4& 5 & 5.6 & 30.6 & \textbf{2.8} \\
        & 6 & \textbf{7.0} & 47.9 & 11.2 \\
        & 7 & \textbf{7.5} & 92.7 & 100.7 \\ \midrule
        Set 5& 5 & 5.8 & \textbf{0.6} & 1.8\\
        & 6 & 7.2 & \textbf{0.8} & 8.6 \\
        & 7 & 9.6 & \textbf{3.1} & 24.1 \\
        \bottomrule
    \end{tabular}
    \caption{Comparison of median execution times in ms. For all G-PCGRL runs, we used models with the graph-wide representations.}
    \label{tab:execution-time}
    \vspace{-4mm}
\end{table}

The random search method operates by randomly selecting edges within the permissible search space between nodes until the specified constraints are satisfied. This approach doesn't involve learning from previous runs or using prior knowledge.
Our implementation of the EA is based on the EA of Rogers et al. to generate graphs to create game economies of varying complexity~\cite{rogers_using_2023}. We also randomly select nodes for crossover, replacing them with their connections between individuals, while mutations occur at a rate of 5\%, randomly replacing a node's connections. An individual's fitness is determined by the cumulative sum of constraints that are not met. The algorithm terminates when a valid graph is found.

The results (Table~\ref{tab:execution-time}) indicate that G-PCGRL exhibits notably superior speed in producing valid graphs with the specified combination of configuration and constraints for sets 1, 2, and 3 across various graph sizes.
For set 4, G-PCGRL only performs better at sizes 6 and 7. For set 5, G-PCGRL is slower than the EA and the random search for the size 5, and slower than the EA for sizes of 6 and 7.

\subsection{Case study: Generating Game Economies and Skill Trees}
\label{sec:demo}

\begin{figure}
     \centering
     \begin{subfigure}[b]{0.235\textwidth}
          \centering
             \includegraphics[width=\textwidth]{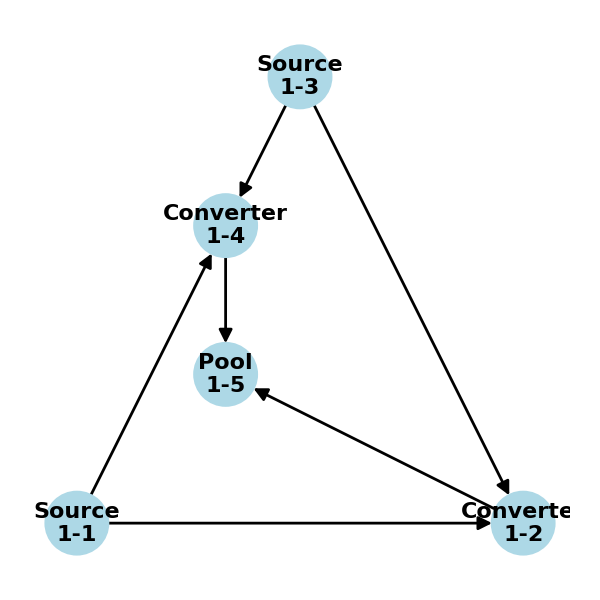}
          \caption{Sample 1 configurations: A graph of size 5 with 2 source nodes, 2 converter nodes, and 1 pool node.}
          \label{fig:sample1}
     \end{subfigure}
     \hspace{0.1cm}
     \begin{subfigure}[b]{0.235\textwidth}
         \centering
         \includegraphics[width=\textwidth]{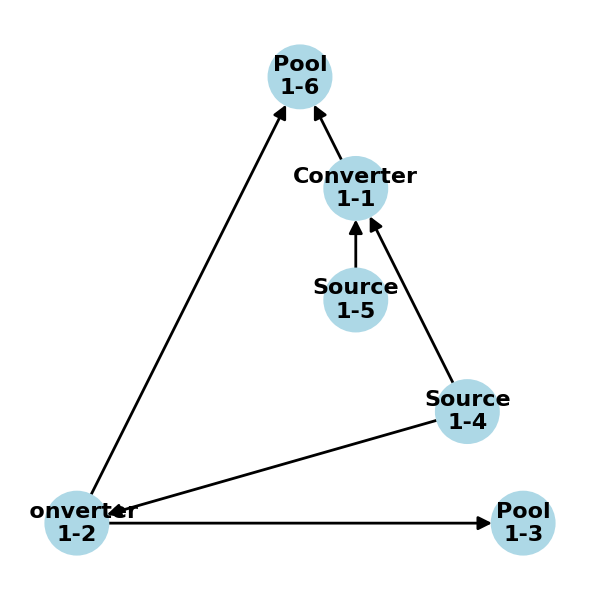}
         \caption{Sample 2 configurations: A graph of size 6 with 3 source nodes, 2 converter nodes, and 1 pool node.}
         \label{fig:sample2}
     \end{subfigure}
  \caption{Generated samples in a game economy setting from one model controlled by different configurations. The configuration controls the graph size and the number of each node type. The model was trained with the set of constraints given in Fig.~\ref{lst-example-constraints}, including three rules and three different node types.}
  \label{fig:samples}
  \vspace{-4mm}
\end{figure}

\begin{figure}[htbp]
     \centering
      \begin{subfigure}[b]{0.45\textwidth}
          \centering
          \vspace{-8mm}
      \includegraphics[width=\textwidth]{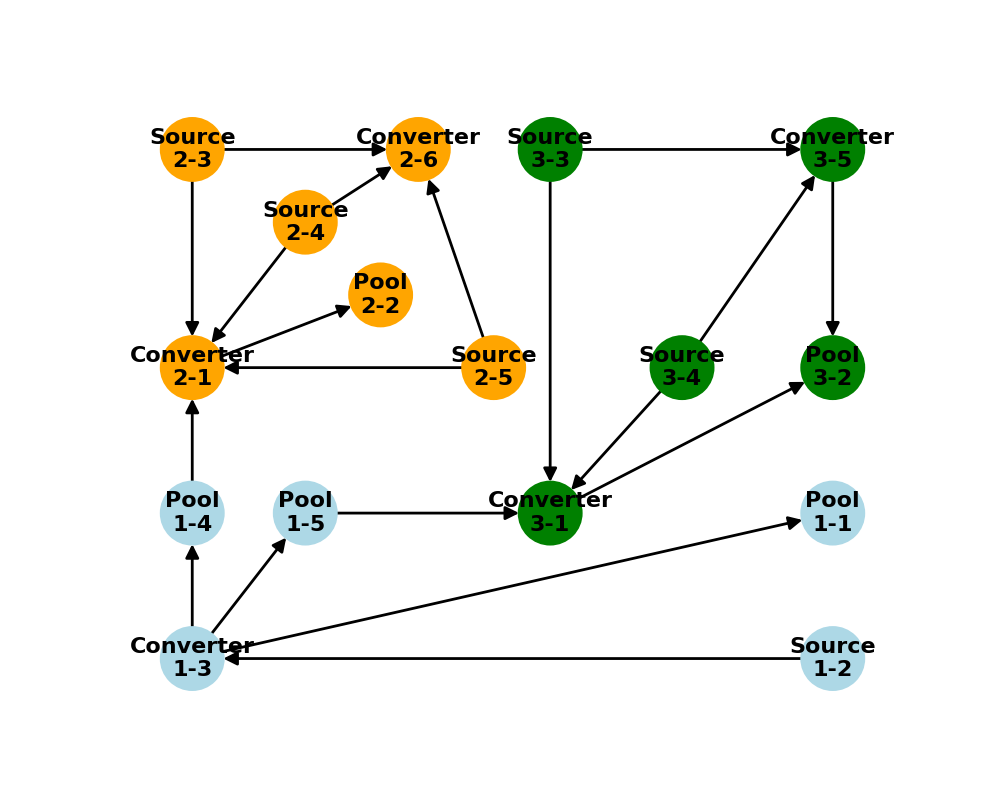}
         
        \vspace{-8mm}
         \caption{An initial economy graph (blue) extended with additional subgraphs (green and orange).}
          \label{fig:concat-economy}
     \end{subfigure}
     \begin{subfigure}[b]{0.41\textwidth}
      \centering
      \includegraphics[width=\textwidth]{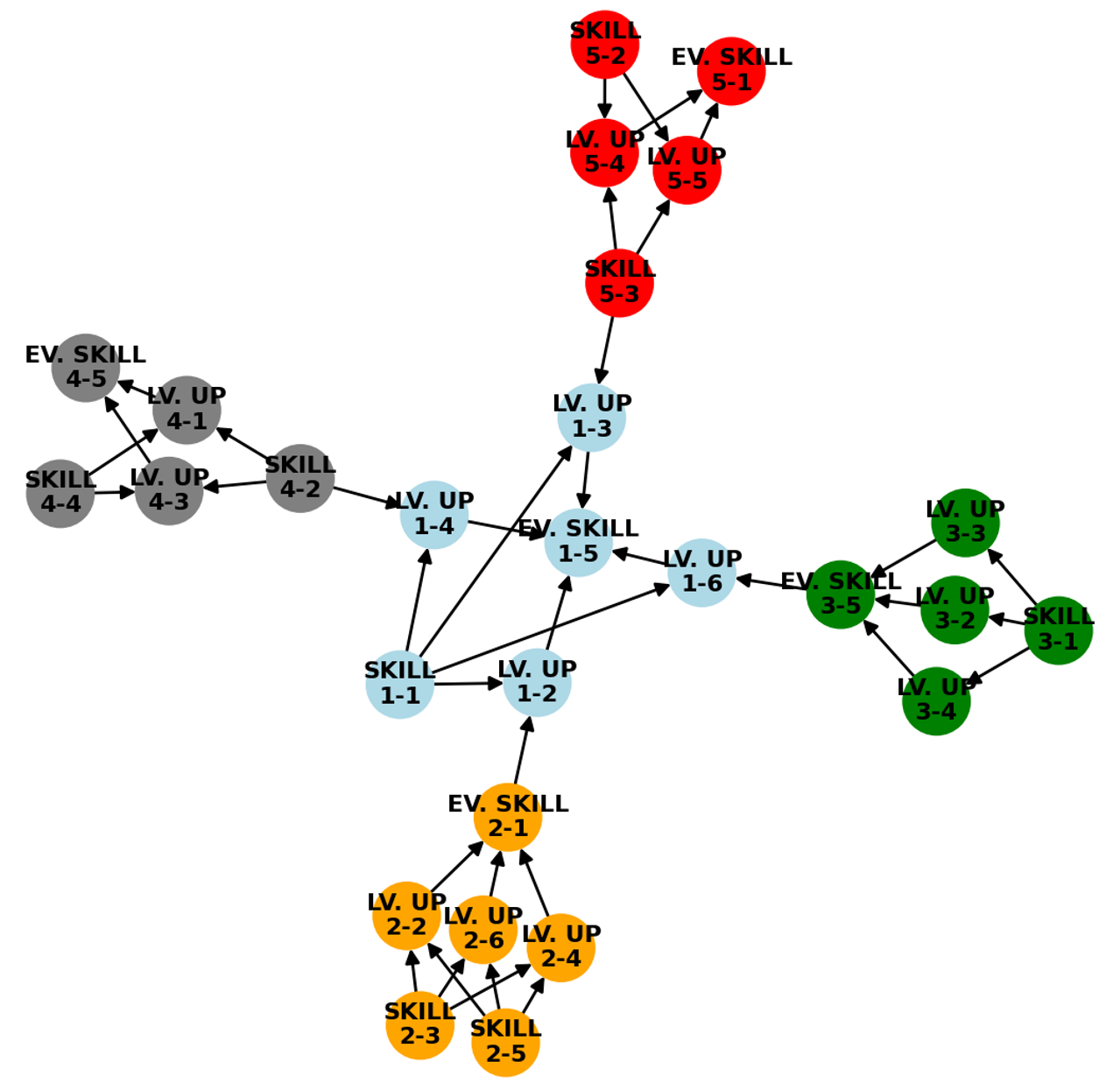}
          \caption{A skill tree built from one initial tree (blue) and concatenated with multiple subgraphs.}
          \label{fig:concat-tree}
     \end{subfigure}
      \caption{The generation of larger graphs by iteratively concatenating the outputs of the same model, each controlled with different configurations. We show this for two domains: game economies (a) and skill trees (b). First, we generate an initial graph (blue). Second, we generate multiple graphs (green, orange, red, and gray), each with different configurations, and concatenate them to the initial graph.
      }
      \label{fig:sample-concat}
      \vspace{-4mm}
\end{figure}

This section provides insights into generated samples in a game economy and skill tree setting. We show how to control a graph-wide model to output different graphs with varying sizes and different numbers of node types.
The model is trained with the set of constraints 1 (Fig.~\ref{lst-example-constraints}) and is controllable for graphs with a maximum size of six with three different node types. A description of the game economy setting is given in Section~\ref{sec:declaration-constraints}.



Fig.~\ref{fig:samples} shows two samples generated from the model with different configurations. Fig. \ref{fig:sample1} shows a graph configured to have a size of five: two source nodes, two converter nodes, and one pool node. In this economy, an item (pool) can be crafted from two source resources, with two conversion processes each requiring both resources. Sample~\ref{fig:sample2} adds another source node, resulting in a graph of size six.
The model now designs the conversion process so that one source is required for both, but the other two sources are required for one each.

\subsubsection{Expanding graphs through concatenation}
\label{sec:concat-skill}

In this section, we explore the possibilities of generating large graphs of arbitrary size by concatenating multiple generated subgraphs. Thus, outputs from the same model, but each controlled with a \emph{different} configuration, increase the diversity of content. The resulting concatenated graph is not a multipartite graph, since there are edges between nodes within the same subgraphs.
For better visualization, we color each subgraph and name the nodes in the schema subgraphIndex-nodeIndex. 
Figure~\ref{fig:sample-concat} shows generated samples of two different domains: an economy graph and a skill tree. For both we use the set of constraints from Fig.~\ref{lst-example-constraints}. First, we generate an initial graph (blue) and expand it with subgraphs. In the context of a game economy~(Fig.~\ref{fig:concat-economy}), items may not be the final product, they may also be the input for another conversion process. Therefore, we connect a randomly selected \texttt{Converter} node to a \texttt{Pool} node from the initial graph. This results in a more complex crafting path for e.g., Pool 3-2, which now also requires Pool 1-5 and Source 1-2. 

In Fig.~\ref{fig:concat-tree} we use G-PCGRL to generate a skill tree from subgraphs. In the context of a skill tree node type $\mathcal{U}$ stands for \texttt{Skill}, $\mathcal{V}$ for \texttt{Lv.up} (level up), and $\mathcal{W}$ for \texttt{Ev. skill} (evolved skill). This implements the abstract concept of having basic skills that can be enhanced by leveling up, leading to the development of more advanced abilities that can be further improved. The leveling path of skills can vary depending on the player's preferences and choices.
For example, a player could start with the basic skill 5-3 and level it up to skill 1-5 or the evolved skill 5-1. The latter could also be achieved by choosing a different path and leveling up skill 5-2 first.

\section{Discussion and Limitations}
\label{sec:discussion}

The results of the experiments with G-PCGRL showed promising results regarding the feasibility of procedurally generating graph-based data from a set of constraints by manipulating the adjacency matrix with RL. There are, however, several things that need to be discussed.

We presented two different representations, graph-narrow and graph-wide, inspired by the respective PCGRL representations. Apart from minor differences, both appear to perform equally well in terms of the proportion of valid graphs generated in a sample of 500.
We trained multiple controllable models with different sets of constraints, node types, and graph sizes. As graph sizes increase, the validity metrics for both representations become worse. This is due to the fact that increasing the maximum graph size nonlinearly increases the complexity and thus the search space for the RL. Furthermore, the models are trained to be controllable \emph{up to} the given size. This also increases the search space and requires more training iterations to learn the controllable configurations. Both are the reasons for the observed results. Training the models to be uncontrollable would improve these results due to the less complex search space. In our opinion, however, it is more beneficial to be able to generate graphs with different configurations using the same model. Therefore, to address the performance loss with increasing graph size, we proposed to concatenate subgraphs generated from the same model but with different configurations.
In comparison, the PCGRL wide baseline performs significantly worse. After all, PCGRL was not designed to generate graph data, but it shows that the graph representations bring improvements to solving this problem.

We compared the execution times of G-PCGRL with an EA and a random search approach across various sets of constraints and graph sizes. Notably, G-PCGRL outperforms both alternative methods in terms of fast content generation, which is particularly evident for the example sets 1, 2, and 3. Unlike the EA and the random search, the execution time of our method is solely dependent on the changes made (cf. Fig.~\ref{fig:performance}, right), showcasing superior robustness compared to the other approaches being strongly dependent on randomness.
Additionally, we examined the execution speed for constraint sets 4 and 5. In particular, set 5 allows for a large number of valid graphs within its search space. Consequently, random-based approaches demonstrate faster execution times for comparably simple constraints compared to our method, primarily due to the wide range of solutions within the search space.

On average, the graph-wide representation requires significantly fewer iterations to generate valid graphs than the graph-narrow~(Fig.~\ref{fig:performance}). This is due to its ability to see the full adjacency matrix and then predict where to add or remove an edge. With this capability, it is possible to create a valid graph more accurately and therefore faster. On the other hand, the complexity of the model and the number of training steps required is increased. However, the faster generation combined with the controllability adds the greatest value and we thus recommend using the graph-wide representation.

We have shown how the model can learn to generate graphs based on a set of constraints and have experimented with different sets and rule combinations. A limitation so far is that only one type of constraint is supported: if specified in a set of a node type, at least one node of a type must be connected. Future work will extend this to implement additional types of constraints, such as explicitly excluding node types or requiring a certain number of edges for a particular node type.
Finally, the reward function motivates the agent to create missing edges and remove wrong ones. This tempts the agent to create graphs with as many edges as possible to increase the reward, potentially reducing the diversity of content created.

In summary, we see this work as a foundational examination, where we first apply PCGRL to the generation of graph data in games in general. We see a high potential for further research in the procedural generation of graph data for games.

\section{Conclusion and Future Work}
\label{sec:conclusion}

We introduced G-PCGRL, a controllable method for procedural content generation (PCG) for graph data from a given set of constraints with reinforcement learning. 
In the early stages of game design, there are usually only abstract concepts. G-PCGRL embeds these concepts as constraints and learns to generate graph-based content accordingly to support designers in the process.
To frame the problem as a Markov decision process, we introduced two new representations for graph data generation for the Procedural Content Generation via Reinforcement Learning (PCGRL) framework.
In addition, our method is controllable in terms of the output to be generated and performs better than the plain PCGRL on this task. Moreover, we demonstrated a superior generation speed for more complex sets of constraints when compared to both an evolutionary algorithm and a random search.
We further showed feasibility through experiments with different representations, graph sizes, sets of constraints, and applied our method to the generation of two game-related domains.
Since the action space does not grow linearly with the maximum controllable graph size, we noticed a drop in performance in terms of the proportion of valid graphs generated for larger sizes. To address this shortcoming in generating larger graphs, we proposed a recursive concatenation of generated graphs from the same model controlled with different configurations.

For future work, we want to experiment with different architectures as feature extractors for the deep reinforcement learning, such as convolutional or graph convolutional layers, in order to increase the scaling of G-PCGRL. In addition, it is interesting to extend the variety of allowed constraint types to allow a more flexible definition of the graphs to be generated.

\bibliographystyle{unsrt}
\bibliography{literature.bib}

\section*{Appendix}

\begin{figure}[htb]
     \begin{subfigure}[b]{0.23\textwidth}
        \centering
        \begin{lstlisting}[frame=tb, language=Python, linewidth=1\textwidth, xleftmargin=0.0\textwidth, xrightmargin=0.0\textwidth,escapechar=§]
      §$\mathcal{U}$§: [§$\mathcal{V}$§]
      §$\mathcal{V}$§: [§$\mathcal{U}$§]
        \end{lstlisting}
          \caption{Set of constraints 2: two rules with two node types.}
          \label{lst:constraints2}
     \end{subfigure}
     \hspace{0.5cm}
     \begin{subfigure}[b]{0.23\textwidth}
         \centering
        \begin{lstlisting}[frame=tb, language=Python, linewidth=1\textwidth, xleftmargin=0.01\textwidth, xrightmargin=0.01\textwidth,escapechar=§]
     §$\mathcal{U}$§: [§$\mathcal{V}$§]
     §$\mathcal{V}$§: [§$\mathcal{U}$§, §$\mathcal{V}$§]
        \end{lstlisting}
          \caption{Set of constraints 3: three rules with two node types.}
         \label{lst:constraints2}
         
     \end{subfigure}   
     \begin{subfigure}[b]{0.23\textwidth}
         \centering
        \begin{lstlisting}[frame=tb, language=Python, linewidth=1\textwidth, xleftmargin=0.00\textwidth, xrightmargin=0.00\textwidth,escapechar=§]
      §$\mathcal{U}$§: §$\hspace{0.35em}$§[§$\mathcal{U}$§]
      §$\mathcal{V}$§: §$\hspace{0.375em}$§[§$\mathcal{U}$§, §$\mathcal{V}$§, §$\mathcal{W}$§]
      §$\mathcal{W}$§: [§$\mathcal{V}$§]
        \end{lstlisting}
          \caption{Set of constraints 4: five rules with three node types.}
         \label{lst:constraints4}
     \end{subfigure}
          \hspace{0.5cm}
     \begin{subfigure}[b]{0.23\textwidth}
         \centering
     \hspace{3cm}
        \begin{lstlisting}[frame=tb, language=Python, linewidth=1\textwidth, xleftmargin=0.01\textwidth, xrightmargin=0.01\textwidth,escapechar=§]
     §$\mathcal{U}$§: §$\hspace{0.35em}$§[§$\mathcal{V}$§, §$\mathcal{W}$§]
     §$\mathcal{V}$§: §$\hspace{0.375em}$§[§$\mathcal{U}$§, §$\mathcal{W}$§]
     §$\mathcal{W}$§: [§$\mathcal{U}$§, §$\mathcal{V}$§]
        \end{lstlisting}
          \caption{Set of constraints 5: six rules with three node types.}
         \label{lst:constraints5}
     \end{subfigure}
     
  \caption{Overview of the sets of constraints used for the experiments in Section~\ref{sec:experiments}. The sets differ in the rules themselves, the number of node types used, and the number of constraints. In total, five sets of constraints were used for the experiments in this work; set 1 is included in Fig.~\ref{lst-example-constraints}.}
  \label{fig:app-constraints}
\end{figure}



\end{document}